# Training Simplification and Model Simplification for Deep Learning: A Minimal Effort Back Propagation Method

Xu Sun, Xuancheng Ren, Shuming Ma, Bingzhen Wei, Wei Li, Jingjing Xu, Houfeng Wang, and Yi Zhang

**Abstract**—We propose a simple yet effective technique to simplify the training and the resulting model of neural networks. In back propagation, only a small subset of the full gradient is computed to update the model parameters. The gradient vectors are sparsified in such a way that only the top-$k$ elements (in terms of magnitude) are kept. As a result, only $k$ rows or columns (depending on the layout) of the weight matrix are modified, leading to a linear reduction in the computational cost. Based on the sparsified gradients, we further simplify the model by eliminating the rows or columns that are seldom updated, which will reduce the computational cost both in the training and decoding, and potentially accelerate decoding in real-world applications. Surprisingly, experimental results demonstrate that most of time we only need to update fewer than 5% of the weights at each back propagation pass. More interestingly, the accuracy of the resulting models is actually improved rather than degraded, and a detailed analysis is given. The model simplification results show that we could adaptively simplify the model which could often be reduced by around 9x, without any loss on accuracy or even with improved accuracy.

✦

## 1 INTRODUCTION

NEURAL network learning is typically slow, where back propagation usually dominates the computational cost during the learning process. Back propagation entails a high computational cost because it needs to compute full gradients and update all model parameters in each learning step ([2], [3], [4]). It is not uncommon for a neural network to have a massive number of model parameters, especially when dealing with large-scale and challenging tasks ([5], [6], [7], [8]).

In this study, we propose a minimal effort back propagation method, which we call *meProp*, for neural network learning. We compute only a very small but critical portion of the gradient information, and update only the corresponding minimal portion of the parameters in each learning step. This leads to sparsified gradients, such that only highly relevant parameters are updated, while other parameters stay untouched. The sparsified back propagation leads to a linear reduction in the computational cost.

On top of meProp, we further propose to simplify the model by eliminating the less relevant neurons discovered during meProp, so that the computational cost of decoding can also be reduced. We name the method *meSimp* (minimal effort simplification). The idea is that we record which neurons are updated at each learning step in meProp, and gradually remove the neurons that are less updated. This leads to a simplified model that costs less in computation

• X. Sun, X. Ren, S. Ma, B. Wei, W. Li, J. Xu, H. Wang, and Y. Zhang are with Institute of Computational Linguistics, School of Electronics Engineering and Computer Science, Peking University. E-mail: {xusun, renxc, shumingma, weibz, liweitj47, ingjingxu, wanghf, zhangyi16}@pku.edu.cn. Corresponding author: Xu Sun.

during decoding, while meProp can only speed up the training of the neural networks.

One of the motivations for such method is that if we suppose the gradients determine the importance of input features, with meProp, the *essential features* are well-trained, and the *non-essential features* are less-trained, so that the model can learn features that are more robustness and overfitting can be reduced. As the essential features play a more important role in the final model, there are chances that the parameters related to non-essential features could be eliminated, which inspires meSimp.

For a classification task, we can categorize the features into: *essential features* that are decisive in the classification, *non-essential features* that are helpful but can also be distractions, and *irrelevant features* that are not useful at all. For example, when classifying a picture as a taxi, the taxi sign is one of the essential features, and the color yellow, which is often the color of a taxi, is one of the non-essential features. Overfitting often occurs when the non-essential features are given too much importance in the model, while meProp intentionally focuses on training the probable essential features to lessen the risk of overfitting.

To realize our approaches, we need to answer four questions. The first is how to find the essential features from the current sample in stochastic learning. We propose a top-$k$ search method based on gradients in back propagation. Interestingly, experimental results demonstrate that most of the time we only need to update fewer than 5% of the parameters, which does not result in a larger number of training iterations. Moreover, this simple strategy can be applied to various neural models, including multi-layer perceptrons (MLP), long-short term memories (LSTM [9]), and convolutional neural networks (CNN [10]).

The second question is whether or not this minimal effort back propagation strategy would hurt the accuracy of the trained models. We show that our strategy does not degrade the accuracy of the trained model, even when a very small portion of the parameters is updated. Interestingly, our experimental results reveal that our strategy actually improves the model accuracy in most cases. Based on our experiments, we find that it is probably because the minimal effort update does not modify weakly relevant parameters in each update, making overfitting less likely.

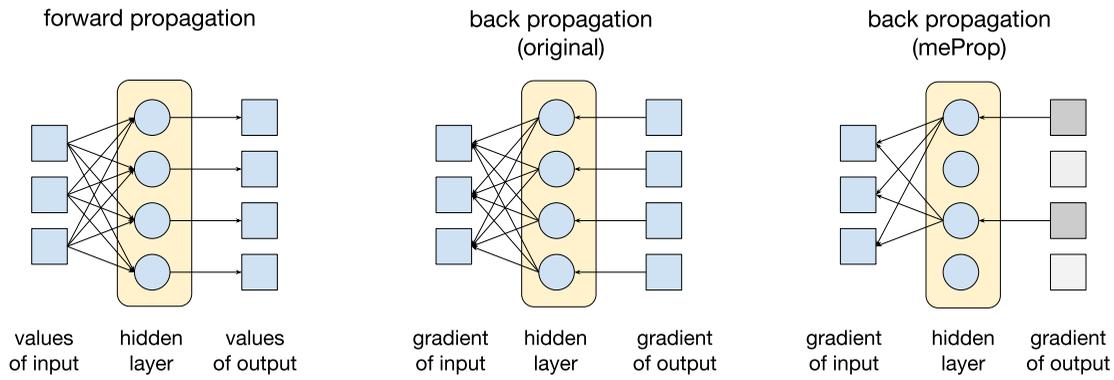

Figure 1. An illustration of training simplification (meProp) with $k = 2$. The forward propagation is computed as usual. During the back propagation, the gradients are sparsified such that only the top-$k$ components in terms of magnitude are kept.

The third question is whether or not the decoding cost of the model can be reduced, as meProp can only shorten the training time. Based on meProp, we further apply the technique of meSimp. From our observations, the simplifying strategy can indeed shrink the final model by usually around 9x without any loss on accuracy. It also supports our assumption that, in fact, many learned features are not essential to the final correct prediction.

The final question is whether or not the size of the model can be determined adaptively. In most previous work, the final model size is pre-configured as desired or using heuristic rules, making it hard to simplify models with multiple layers, because naturally, each layer should have a different dimension, since it captures a different level of abstraction. In practice, we find that meSimp could adaptively reduce the size of the hidden layers, and automatically decide which features are essential for the task at different abstraction levels, resulting in a model of different hidden layer sizes.

The contributions of this work are as follows:
- We propose a minimal effort back propagation technique for neural network learning, which automatically finds the most important features based on the magnitude of the gradients. This method works for different types of deep learning models (MLP, CNN, and LSTM) and diverse tasks (natural language processing and image recognition).
- Applying the technique to training simplification (meProp), we find that the strategy actually improve the accuracy of the resulting models, rather than degraded, even if fewer than 5% of the weights are updated at each back propagation pass most of the time. The technique does not entail a larger number of training iterations, and could reduce the time of the training substantially.
- Most importantly, applying the technique to model simplification (meSimp) could potentially reduce the time of decoding. With the ability to adaptively simplify each layer of the model to only keep essential features, the resulting model could be reduced to around one ninth of its original size, which equals to an around 9x reduction in decoding cost, on a base of no accuracy loss or even improved accuracy. It's worth mentioning, when applied to models with multiple layers, given a single hyper-parameter, meSimp could simplify each hidden layer to a different extent, alleviating the need to set different hyper-parameters for different layers.

## 2 PROPOSED APPROACH

We propose a simple yet effective technique for neural network learning. The forward propagation is computed as usual. During back propagation, only a small subset of the full gradients are computed to update the model parameters. The gradients are sparsified so that only the top-$k$ components in terms of magnitude are kept. Based on the technique, we further propose to simplify the resulting models by removing the neurons that are seldom updated, according to the top-$k$ indices. The model is simplified in such a way that only actively updated neurons are kept. We first present the proposed methods, and then describe the implementation details.

### 2.1 Simplified Back Propagation (meProp)

Forward propagation of neural network models consists of linear transformations and non-linear transformations. For simplicity, we take a computation unit with one linear transformation and one non-linear transformation as an example:

$$\boldsymbol{y} = W\boldsymbol{x} \tag{1}$$
$$\boldsymbol{z} = \sigma(\boldsymbol{y}) \tag{2}$$

where $W \in \mathbb{R}^{n \times m}, \boldsymbol{x} \in \mathbb{R}^m, \boldsymbol{y} \in \mathbb{R}^n, \boldsymbol{z} \in \mathbb{R}^n$, $m$ is the size of the input vector, $n$ is the size of the output vector, and $\sigma$ is a non-linear function (e.g., $ReLU$, and $tanh$). We can consider that each row of $W$ represents a neuron that transforms an input to an activation. During back propagation, we need to

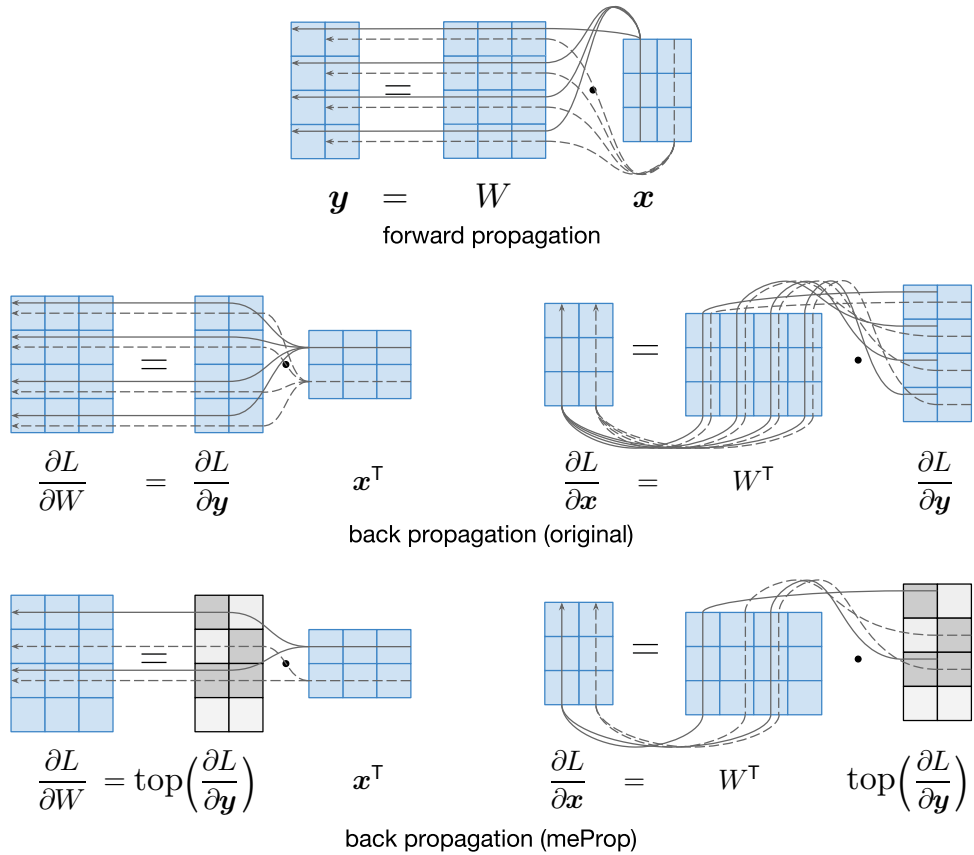

Figure 2. An illustration of the computational flow of meProp in the mini-batch scenario. Here, the mini-batch size is 2, and $k = 2$. Each row in $W$ represents a neuron. Examples are represented as column vectors. The sold lines and the dashed lines show the computation flow of the two examples. Compared with original back propagation, meProp only back propagates from the output with the largest gradients, which are shown in darker grey and may be different for each example. The procedure involves sparse computations and linear reduction in computational complexity. As we illustrate, in back propagation, there may be neurons that are not used for any example in the mini-batch (the last neuron in this case).

compute the gradient w.r.t. the parameter matrix $W$ and the input vector $x$:

$$\left(\frac{\partial z}{\partial W}\right)_{ij} = \sigma'_i x_j^\top \quad (1 \leq i \leq n, 1 \leq j \leq m) \quad (3)$$

$$\left(\frac{\partial z}{\partial x}\right)_i = \sum_j W_{ij}^\top \sigma'_j \quad (1 \leq j \leq n, 1 \leq i \leq m) \quad (4)$$

where $\sigma'_i$ means $\partial z/\partial y_i$. We can see that the computational cost of back propagation is directly proportional to the size of output vector $n$.

The proposed meProp uses approximate gradients by keeping only top-$k$ elements based on the *magnitude values*. That is, only the top-$k$ elements with the largest absolute values are kept. For example, suppose a vector $v = \langle 1, 2, 3, -4 \rangle$, then $\text{top}_2(v) = \langle 0, 0, 3, -4 \rangle$. We denote the indices of vector $\sigma'$'s top-$k$ values as $S = \{t_1, t_2, ..., t_k\}(1 \leq k \leq n)$, and the approximate gradient of the parameter matrix $W$ and input vector $x$ is:

$$\left(\frac{\partial z}{\partial W}\right)_{ij} \leftarrow \sigma'_i x_j^\top \quad \text{if } i \in S \text{ else } 0 \quad (5)$$

$$\left(\frac{\partial z}{\partial x}\right)_i \leftarrow \sum_j W_{ij}^\top \sigma'_j \quad \text{if } j \in S \text{ else } 0 \quad (6)$$

As a result, only $k$ rows of the weight matrix are modified, leading to a linear reduction ($k$ divided by the vector dimension) in the computational cost. The algorithm is described in Algorithm 1.

Figure 1 is an illustration of meProp for a single neuron. The original back propagation uses the full gradient of the output vectors to compute the gradient of the parameters. The proposed method selects the top-$k$ values of the gradient of the output vector, and back propagates the loss through the corresponding subset of the parameters.

---

**Algorithm 1** Back propagation simplification for a computation unit

**Input:** $y \leftarrow Wx, z \leftarrow \sigma(y)$
1: $\sigma' \leftarrow \partial z/\partial y$ ▷ Gradient of $y$ w.r.t. $z$
2: $S \leftarrow \{t_1, t_2, ..., t_k\}$ ▷ Indices of $k$ largest derivatives of $\sigma'$ in magnitude
3: $(\partial z/\partial W)_{ij} \leftarrow \sigma'_i x_j^\top$ if $i \in S$ else $0$
4: $(\partial z/\partial x)_i \leftarrow \sum_j W_{ij}^\top \sigma'_j$ if $j \in S$ else $0$

---

As for a complete neural network framework with a loss $L$, the original back propagation computes the gradient of the parameter matrix $W$ as:

$$\frac{\partial L}{\partial W} = \frac{\partial L}{\partial y} \frac{\partial y}{\partial W} \quad (7)$$

while the gradient of the input vector $x$ is:

$$\frac{\partial L}{\partial x} = \frac{\partial y}{\partial x}\frac{\partial L}{\partial y} \qquad (8)$$

The proposed meProp selects top-$k$ elements of the gradient $\partial L/\partial y$ to approximate the original gradient, and passes them through the gradient computation graph according to the chain rule. Hence, the gradient of $W$ goes to:

$$\frac{\partial L}{\partial W} \leftarrow \text{top}_k\left(\frac{\partial L}{\partial y}\right)\frac{\partial y}{\partial W} \qquad (9)$$

while the gradient of the vector $x$ is:

$$\frac{\partial L}{\partial x} = \frac{\partial y}{\partial x}\text{top}_k\left(\frac{\partial L}{\partial y}\right) \qquad (10)$$

Figure 2 shows an illustration of the computational flow of meProp. The forward propagation is the same as traditional forward propagation, which computes the output vector via a matrix multiplication operation between two input tensors. The original back propagation computes the full gradient for the input vector and the weight matrix. For meProp, back propagation computes an approximate gradient by keeping top-$k$ values of the backward flowed gradient and masking the remaining values to 0.

## 2.2 Simplified Model (meSimp)

The method from Section 2.1 simplifies the training process, and thus can reduce the training time. However, for most deep learning applications in real life, it is even more important to reduce the computational cost of decoding because decoding needs to be done as long as there is a new request. Although training can be time consuming, it only needs to be done once.

In this section, we propose to simplify the model by eliminating the *inactive paths*, which we define as the neurons whose gradients are not in top-$k$. This way, the decoding cost would also be reduced. There are two major concerns about this proposal. The main problem here is that we do not know the active path of unseen examples in advance, as we do not know the gradient information of those examples. Our solution for this problem is that we could obtain the overall inactive paths from the inactive paths of the training examples, which could be removed gradually in the training. The second is that the reduction in dimension could lead to performance degradation. Surprisingly, from our experimental results, our top-$k$ gradient based method does not deteriorate the model. Instead, with an appropriate configuration, the resulting smaller model often performs better than the baseline large model, or even the baseline model of the similar size. As a matter of fact, after pruning the performance does drop. However, with the following training, the performance is regained. In what follows, we will briefly introduce the inspiration of the proposed method, and how the model simplification is done.

In the experiments of meProp, we discover an interesting phenomenon that during training, apart from the active paths with top-$k$ gradients, there are some inactive paths that are not activated at all for any of the examples. We call these paths *global inactive paths*. These neurons are not updated at all during training, and their parameter values remain the same as their initialized values, and we have every reason to believe that they would not be effective for new samples as well. However, the number of those paths is not sufficient to bring a substantial contraction to the model.

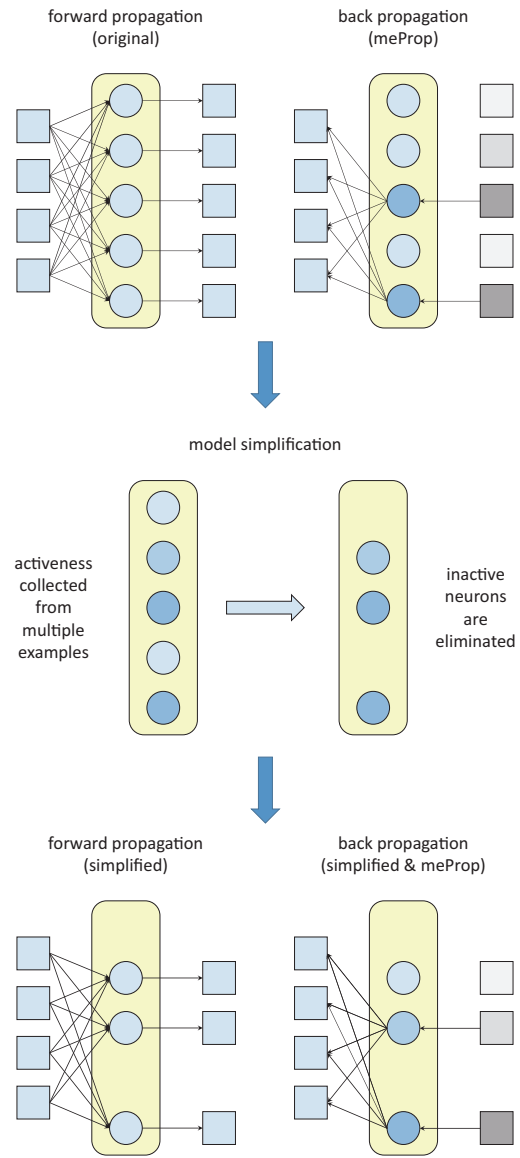

Figure 3. An illustration of model simplification (meSimp) with $k = 2$. The figure shows the three main stages of the simplified model training. First, the model is trained using meProp for several iterations and a record of the activeness of the paths is kept, indicated by the shades of the neurons. Second, the model is simplified based on the collected record and the inactive paths are eliminated. Third, the simplified model is trained again using meProp. We repeat the procedure until the goal of the training is met.

Based on the previous findings, we generalize the idea of universal inactive paths, and prune the paths that are less updated, that is, the paths we eliminate are the paths that are not active for a number of samples. To realize the idea, we keep a record $c$ of how many times the index is in the top-$k$ indices $S$, during the back propagation at the same time as meProp. After several training steps, we take out the less active paths that are not updated for a number of samples, e.g. 90%, which results in a simplified model.

The record is cleared at each pruning action. By doing that iteratively, the model size will approach near stable in the end. Algorithm 2 describes the method for a computation unit, and an illustration is shown in Figure 3.

An important hyper-parameter for the method is the pruning threshold. When determining the threshold, the model size and the number of examples between pruning actions should be taken into account. As shown in Algorithm 2, the threshold could be parameterized by prune interval $m$, which is the number of examples between pruning, and prune rate $p$, which represents how active the path should be if it is not to be eliminated.

---

**Algorithm 2** Model simplification for a computation unit

1: Initialize $W$, $t \leftarrow 0$, $\boldsymbol{c} \leftarrow \boldsymbol{0}$    ▷ $t$ is the current training step, $\boldsymbol{c}$ is top-$k$ history
2: **while** training **do**
3:    Draw $(\boldsymbol{x}, \boldsymbol{y})$ from training data
4:    $\boldsymbol{z} \leftarrow \sigma(W\boldsymbol{x})$    ▷ Forward propagation
5:    $\frac{\partial \boldsymbol{z}}{\partial W} \leftarrow \text{top}_k(\boldsymbol{\sigma}')\boldsymbol{x}^{\mathsf{T}}$    ▷ Gradient of $W$ w.r.t. $\boldsymbol{z}$
6:    $S \leftarrow \{t_1, t_2, ..., t_k\}$    ▷ indices of $k$ largest derivatives of $\boldsymbol{\sigma}'$ in magnitude
7:    $c_i \leftarrow$ increase if $i$ in $S$    ▷ Record the top-$k$ indices
8:    Update $W$ with $\frac{\partial \boldsymbol{z}}{\partial W}$
9:    **if** $t \bmod m = 0$ **then**    ▷ Prune inactive paths
10:       $\theta \leftarrow m \times p$
11:       **for all** $i$ where $c_i < \theta$ **do**
12:          Remove row $i$ from $W$
13:       **end for**
14:       $\boldsymbol{c} \leftarrow \boldsymbol{0}$    ▷ Reset $\boldsymbol{c}$
15:    **end if**
16:    $t \leftarrow t + 1$
17: **end while**

---

Note that the layer sizes are determined adaptively in a multi-layer setting, and only one pruning rate $p$ is needed to obtain different layer sizes. Because the top-$k$ indices of different layers at different iterations intersect differently in back propagation. For some layers, the top-$k$ indices are similar, hence results in a larger layer size, compared to $k$. For other layers, the top-$k$ indices are quite different at each iteration, so that the intersection happens more often, which means $c_i$ is lower, hence the resulting layer size is smaller. How the layer is simplified depends on how the learning is done, which is in accordance with our intuition.

In a deep neural network, it is worth noticing that when simplifying a hidden layer, the respective columns in the following layer could also be removed, as the values in the columns represent the connection between the eliminated inputs and the outputs, which is no longer effective. That could reduce the model even further.

**Cycled Training.** For model simplification, we also propose a kind of cycle mechanism. During our experiments, we find that at the time of the simplification, there is a drop in performance, but it recovers quickly within the following training, and may even supersede the performance before the simplification. It makes us wonder whether the training after simplification is critical to the performance improvement. We propose to divide the training procedure into several stages, and in each stage, we first conduct the training of model simplification, and then conduct the normal training. At the start of each stage, we also reinitialize the optimizer, if there is historical information of the gradients stored. The reason for such operation is that after model simplification, the dynamics of how the neurons interacted with each other changed, and the previous gradient information may interfere with the new dynamics of the simplified network. We find this cycle mechanism could improve the resulting model's performance even further on some tasks.

### 2.3 Implementation

Most of the experiments on CPU are conducted on the framework coded in C# on our own, which does not make use of matrix optimization techniques to illustrate the theoretical effect on computational complexity. We also have an implementation based on the PyTorch framework for GPU based experiments.

#### 2.3.1 Where to Apply Top-k Selection

The proposed method aims to reduce the complexity of the back propagation by reducing the elements in the computationally intensive operations. In our preliminary observations, matrix-matrix or matrix-vector multiplication consumed more than 90% of the time of back propagation in a fully-connected layer. Therefore, the proposed approach is applied only to the back propagation of matrix multiplications. For other element-wise operations (e.g., activation functions, and additive functions), the original back propagation procedure is kept, because those operations are already fast enough compared with matrix multiplications. As a reminder, since the gradient of the input of a layer is still dense through the sparsified back propagation, the top-$k$ selection needs to be applied to every layer.

For many neural models, the fully-connected layer serve as the final output layer and directly receives the error signal from the loss. Its size is typically determined by the task and can be quite different from the hidden layers. When its size is very small, e.g., 10 for MNIST, it is not rational to apply the proposed approach, because the computations saved are negligible and the source error signal may be perturbed. For other tasks, e.g., Parsing, where the output layer size is larger, the proposed approach may be applied, but the best $k$ may be different from the hidden layers.

#### 2.3.2 How to Apply Top-k Selection to Recurrent Layers

While the implementation for vanilla recurrent layers is straight-forward, in practice, many other advanced recurrent layers are used, such as LSTMs and the Gated Recurrent Units (GRU [11]). They are based on the gate mechanism, where multiple gates control how the output is generated. It is natural to ask whether the gates should be considered independently or jointly when applying the proposed approach, because the corresponding neuron in each gate controls the same output feature. In our preliminary experiments, we find that the difference is minimal, so we simplified the back propagation of the gates independently, which can better capture the behaviour of each gate. However, when simplifying the layer, it makes sense only if the gates are simplified jointly. Otherwise, the simplification would cause each gate control a different set of output features.

Table 1
Sizes of Datasets in Main Experiments. For Parsing and POS, we list the number of sentences first, and for MNIST, images. The number in parentheses means the number of actual classifications, which is the number of words for POS and the number of transitions for Parsing.

| Dataset | Train | Dev | Test |
|---|---|---|---|
| Parsing | 39,832 (1,900,056) | 1,700 (80,234) | 2,416 (113,368) |
| POS | 38,219 (912,344) | 5,527 (131,768) | 5,462 (129,654) |
| MNIST | 55,000 (55,000) | 5,000 (5,000) | 10,000 (10,000) |

Table 2
Experimental Results of meProp. Typically, fewer than 5% of the weights are sufficient for meProp to reach the same performance (best of 5 runs) with substantial speedups in back propagation.

| Parsing | Iter | Backprop time (s) | Dev (%) | Test (%) |
|---|---|---|---|---|
| MLP (h=500) | 10 | 9,077.7 | 90.48 | 89.80 |
| meProp (k=20) | 6 | 488.7 (18.6x) | 89.91 | 89.84 (+0.04) |
| **POS** | Iter | Backprop time (s) | Dev (%) | Test (%) |
| LSTM (h=500) | 3 | 16,167.3 | 97.20 | 97.22 |
| meProp (k=10) | 4 | 435.6 (37.1x) | 97.14 | 97.25 (+0.03) |
| **MNIST** | Iter | Backprop time (s) | Dev (%) | Test (%) |
| MLP (h=500) | 13 | 169.5 | 98.72 | 98.20 |
| meProp (k=80) | 14 | 28.7 (5.9x) | 98.36 | 98.27 (+0.07) |

Hence, we decide to simplify the gates jointly. In practice, we simplify the layer after merging the active paths from each gate, so that the most active path across all gates are kept.

## 3 EXPERIMENTS

We perform experiments on various neural models on diverse tasks. Our main experiments of meProp and meSimp are conducted on the following three tasks, the statistics of which are shown in Table 1:

**Transition-based Dependency Parsing (Parsing).** Following [12], we use an MLP with two layers in total as the baseline, except that the hidden size is set to 500 and we do not use dropout. We use the standard splits of the English Penn Treebank (PTB) [13] for this task. The number of transitions[1] are shown in parentheses in Table 1. The evaluation metric is unlabeled attachment score (UAS).

**Part-of-Speech Tagging (POS).** We use an LSTM model with one hidden layer as the baseline. We use the bidirectional version ([14], [15]) to better represent the context. The hidden size is 500 for each direction, thus 1000 in total. We use the standard splits [16] of PTB for this task. The number of words are shown in parentheses in Table 1. The evaluation metric is per-word accuracy.

**MNIST.** MNIST [10] is commonly used for training image processing systems. We use an MLP with three layers in total as the baseline. The hidden size is 500 and we use ReLU ([17], [18]) as the activation function. We select the first 5000 images of the training images as the development set. The evaluation metric is accuracy.

### 3.1 Experimental Settings

Based on the development set and prior work, we set the mini-batch size to 10,000 (transitions) [12], 1 (sentence), and 10 (images) for Parsing, POS, and MNIST, respectively. For all experiments, we use the Adam optimizer with the suggested hyper-parameters in [19] to avoid extensive tuning. Other optimizers with adaptive learning rates, e.g. AdaGrad [20] may also work, as shown in our previous work [1].

In the meSimp experiments, we only simplify the hidden layers of the model. We set the cycle to 10 for all the tasks, that is, we first train the model using meSimp for 5 epochs, then train the model normally for 5 epochs, and repeat the procedure till the end.

To simulate the real-world scenario, we run each configuration 5 times with different random seeds in the main experiments, and choose the best model with the minimum $k$ on the development set to report, although bigger $k$ may result in better results. The time is measured on a computer with an *Intel Xeon 3.0GHz CPU* and a *NVIDIA GeForce GTX 1080 GPU*.

### 3.2 Experimental Results of meProp

Table 2 shows the results based on different models. In the table, *meProp* means applying meProp to the corresponding baseline model, $h = 500$ means that the hidden layer dimension is 500, and $k = 20$ means that meProp selects top-20 elements in back propagation. We report the minimum $k$ that results in comparable performance under the settings in Section 3.1.

Since meProp is applied to the linear transformations (which entail the major computational cost), we report the linear transformation related backprop time as *Backprop Time*. It does not include non-linear activations, which usually have only less than 2% computational cost. For the total time of back propagation including non-linear activations, please refer to our previous work [1]. As we can see, applying meProp can substantially speed up the back propagation. It provides a linear reduction in the computational cost. Surprisingly, results demonstrate that we can update only fewer than 5% of the weights at each back propagation pass for the natural language processing tasks. This does not result in a larger number of training iterations.

More surprisingly, the accuracy of the resulting models is actually improved rather than decreased. The reason could be that the neurons that are not important for correct predictions are not considered in meProp, which makes the model effectively smaller so that overfitting is less likely. To some extent, it is similar to dropout [21], where the effective model is made smaller for each example. Moreover, as the forward propagation is kept in meProp, the output value distribution is accurate w.r.t. the full model, while in dropout the output value distribution needs extra scaling to approximate the real distribution w.rt. the full model.

#### 3.2.1 Analysis of meProp

We conduct analysis on MNIST unless otherwise stated, since it is widely used in testing machine learning algorithms [10], [19], [22]. In all the following analysis, we conduct the experiments once but keep the random seed the same and fixed to facilitate fair comparisons.

---

1. In transition-based systems, a training example consists of a parsing context and its optimal transition action.

Table 3
meProp: Results Based on the Same $k$ and $h$. It suggests meProp can learn better and more robust features, since meProp outperforms both the original big baselines and the baselines with hidden layers of size $k$.

| Parsing | Iter | Test (%) |
|---|---|---|
| MLP (h=20) | 18 | 88.37 |
| meProp (k=20) | 6 | **90.01 (+1.64)** |
| **POS** | Iter | Test (%) |
| LSTM (h=5) | 7 | 96.40 |
| meProp (k=5) | 5 | **97.12 (+0.72)** |
| **MNIST** | Iter | Test (%) |
| MLP (h=20) | 15 | 95.77 |
| meProp (k=20) | 17 | **98.01 (+2.24)** |

Table 4
meProp: Results with Dropout. It can be concluded that meProp can further improve the performance with dropout.

| Parsing | Dropout | Test (%) |
|---|---|---|
| MLP (h=500) | 0.5 | 91.53 |
| meProp (k=40) | 0.5 | **91.99 (+0.46)** |
| **POS** | Dropout | Test (%) |
| LSTM (h=500) | 0.5 | 97.20 |
| meProp (k=20) | 0.5 | **97.31 (+0.11)** |
| **MNIST** | Dropout | Test (%) |
| MLP (h=500) | 0.2 | 98.09 |
| meProp (k=25) | 0.2 | **98.32 (+0.23)** |

Table 5
meProp: Results of More Hidden Layers. As suggested by the results, meProp can also work on deeper models.

| MNIST Test (%) | 2 | 3 | 4 | 5 |
|---|---|---|---|---|
| MLP (h=500) | 98.10 | 98.21 | 98.10 | 98.05 |
| meProp (k=25) | **98.20** | **98.37** | **98.15** | **98.21** |
| Improvement | +0.10 | +0.16 | +0.05 | +0.16 |

Table 6
Unified meProp: Results of *Simple Unified Top-k* meProp. The unified meProp can also achieve good results and work for deeper models.

| MNIST | Test (%) |
|---|---|
| MLP (2 hidden layers, h=500) | 97.97 |
| meProp (k=30) | **98.08 (+0.11)** |
| MLP (5 hidden layers, h=500) | 98.00 |
| meProp (k=50) | **98.09 (+0.09)** |

**Varying Hidden Sizes.** An important question is: does meProp work well simply because the tasks do not require large hidden layers which cause overfitting? If so, we could simply train a smaller network and the performance could be better that both the baseline and the baseline with meProp. To examine this, we perform experiments using the same hidden dimension as $k$, and the results are shown in Table 3. As we can see, however, the results of the small hidden dimensions are much worse than those of meProp. As the baseline and the model of size $k$ always perform worse, we can conclude that the tasks require hidden layers larger than $k$ but smaller than the baseline and meProp does not always select the same $k$ features. Instead, it selects from a large pool of features and chooses the most important features to update, which may be different at each pass.

**Using Dropout.** Since meProp shows the ability to reduce overfitting of the baselines and intuitively resembles the dropout procedure, a natural question is whether meProp works the same as dropout or reduces overfitting in a different way. We choose the dropout rate based on the development sets and then further apply meProp to check if further improvement is possible. Table 4 shows that meProp can achieve further improvement over dropout. In particular, meProp has an improvement of 0.46 on Parsing. The results suggest that the type of overfitting that meProp reduces is probably different from that of dropout. Thus, a model should be able to take advantage of both meProp and dropout to reduce overfitting and learn more robust features.

**Using More Hidden Layers.** It is also important to examine how the top-$k$ selection affects the deeper models. We conduct experiments on MNIST using 2 to 5 hidden layers (excluding the output layer). We set the dropout rate to 0.1, because we find that the baselines with more layers require its use to alleviate overfitting and 0.1 works for most of the layer settings. Table 5 shows that adding the number of hidden layers does not hurt the performance of meProp.

### 3.3 Extending meProp

In this section, we explore extending meProp to achieve speedups on GPUs, to more complex models, and to more challenging tasks.

#### 3.3.1 Acceleration on GPUs

**Unified meProp.** For implementing meProp on GPU, the simplest solution is to treat the entire mini-batch as a "big training example", where the top-$k$ operation is based on the averaged value of each neuron w.r.t. all examples in the mini-batch. In this way, the big sparse matrix of the mini-batch will have consistent sparse patterns among examples, and then can be transformed into a small dense matrix by removing the zero values. We call this implementation Unified meProp, short for Simple Unified Top-$k$ meProp. Despite its simplicity, Table 6 shows good performance, and the implementation also works for deeper models. The size of mini-batch is 50 in the experiments.

We also find the speedup on GPU is less significant when the hidden size is small. The reason is that our GPU's computational power is not fully consumed by the baseline with small hidden layers, so that the original backprop is already fast enough, making it hard for meProp to achieve substantial speedup. For example, supposing a GPU can finish 1000 operations in one cycle, there could be no speed difference between a method with 100 and a method with 10 operations. Indeed, we find MLP (h=64) and MLP (h=512) have almost the same GPU speed (572ms vs. 644ms) even on forward propagation, while theoretically there should be an 8x difference. This provides evidence for our hypothesis that our GPU is not fully consumed with small hidden layers. It is also confirmed by the pioneering work on using GPU for machine learning [23].

Thus, the speedup test on GPU is more meaningful for the heavy models so that the baseline can at least fully consume the GPU's computational power. To verify this, we first test the GPU speedup of a single large computation unit on synthetic data, and Table 7 shows that meProp

Table 7
Unified meProp on GPUs: Speedups of a Computation Unit on Synthetic Data and an MLP with Two Hidden Layers on MNIST. As we can see, the speedups are substantial for heavy models.

| Overall Backprop time (ms) | Synthetic | MNIST |
| --- | --- | --- |
| Baseline (h=8192) | 308.0 | 17,696.2 |
| meProp (k=32) | 11.2 (27.5x) | 1,656.9 (10.7x) |
| meProp (k=64) | 14.4 (21.4x) | 1,828.3 ( 9.7x) |
| meProp (k=128) | 21.3 (14.5x) | 2,200.0 ( 8.0x) |
| meProp (k=256) | 38.6 ( 8.0x) | 3,149.6 ( 5.6x) |
| meProp (k=512) | 70.0 ( 4.4x) | 4,874.1 ( 3.6x) |

Table 8
meProp: Results of CNNs. The results show that it is possible to apply meProp to CNNs and the top-$k$ ratio can be as small as 5% to reach comparable or better performance.

| MNIST | Test (%) |
| --- | --- |
| CNN | 99.37 |
| meProp (k ratio=0.01) | 99.06 |
| meProp (k ratio=0.02) | 99.23 |
| meProp (k ratio=0.05) | 99.38 |
| meProp (k ratio=0.10) | 99.39 |

achieves much higher speed than the original backprop. Then, we test the GPU speedup of an MLP with two large hidden layers [24], and Table 7 shows that meProp also has substantial GPU speedup on MNIST using large hidden layers. The size of mini-batch is 1024 in the experiments. The speedup is based on *Overall Backprop Time*, which is the time of the PyTorch `backward` call. Those results demonstrate that meProp can achieve good speedup on GPU when it is applied to heavy models.

Finally, there are potentially other kinds of implementation of meProp on GPU. For example, another natural solution is to directly use the sparse BLAS procedures, such that the optimized sparse matrix multiplication library is used to accelerate the computation. This could be an interesting direction of future work.

### 3.3.2 Applying to CNNs

The most distinguishable attribute of a convolutional layer [10] from a plain fully-connected layer is that it is based on the convolution operation, or, more precisely, the cross-correlation operation. The weight (or the kernel or the filter) is repeatedly applied to the sub-region of the input, which induces intensive weight sharing and high computational complexity. Fortunately, the convolution operation can be resolved to matrix multiplication with input manipulation [25] so that the method mentioned in Section 2.1 also works. Since most convolutional layers deal with image-related tasks, we focus on the spatial convolution in the following, the input of which typically has three dimensions.

**Spatial Top-$k$ Selection.** Suppose the three dimensions represent the channel, the height, and the width, respectively. There are several ways to conduct top-$k$ selection, and we choose to select top-$k$ from the spatial dimensions, i.e., from each feature map, analogous to spatial pooling operations [26]. The reason is that each output channel is supposed to capture a certain type of visual features, and by selecting top-$k$ gradients from the spatial dimensions, we are back propagating only through the most influential

Table 9
meProp: Results of Sequence-to-Sequence Models. On the challenging machine translation tasks with large-scale data and deep complex models, meProp can also achieve promising results, using 6.25% of the parameters, showing its potential and generality.

| **Zh-En** | Dev BLEU | Test BLEU |
| --- | --- | --- |
| LSTM (h=512) | 39.30 | 36.33 |
| meProp(k=32) | 39.48 | **36.42 (+0.09)** |
| **En-Vi** | Dev BLEU | Test BLEU |
| LSTM (h=512) | 26.62 | 29.33 |
| meProp(k=32) | 26.27 | **29.54 (+0.21)** |

regions. The theoretical computation reduction is also linear to the inverse of $k$.

We conduct experiments on MNIST to verify the proposed strategy. The baseline consists of two $5 \times 5$ convolutional layers with 16 and 32 output channels, each followed by a $2 \times 2$ max pooling layer. Two fully-connected layers of size 512 and 10 are used to make predictions [27]. The activation function is also ReLU. We refrain from using other techniques, such as dropout and batch normalization [28], to isolate the effect of meProp on CNNs. As the size of the output features is changing in each layer, we use top-$k$ ratio instead of fixed $k$. We run the experiments 10 times and report the best result of each configuration. The results are shown in Table 8. As we can see, 5% of the full gradients are sufficient to achieve comparable performance. As we use max pooling layers with a $2 \times 2$ receptive region, only 25% of the gradients to convolutional layers are non-zero. Despite that, 80% of the remaining gradients are also not necessary as shown by our results.

### 3.3.3 Applying to Deep Sequence-to-Sequence Models

Sequence-to-sequence models [5] are mainly used in challenging tasks involving natural language generation, e.g., machine translation, which requires large-scale datasets to train. There is an encoder that encodes the input sequence into a fixed-length representation and a decoder that generates the output sequence according to the representation. In general, both the encoder and the decoder are based on recurrent layers, and the attention mechanism [6], [29] may be used to improve the alignments between the output and the input.

**Alternating Top-$k$ Selection.** The main problem of applying meProp to the sequence-to-sequence model is that the dependency length from an output word to an input word can be too long. In practice, we find that directly applying meProp to both the encoder and the decoder at the same time can work, but the results can be unstable. Instead, we propose to applying the top-$k$ sparsification to the encoder and the decoder iteratively. In each epoch, either the encoder or the decoder uses meProp, which can allow the error signal to flow the other part uninterruptedly, while maintaining satisfying reduction in computational complexity.

We conduct experiments on two large-scale machine translation datasets using deep LSTM-based sequence-to-sequence models. The datasets are as follows:
**Chinese-English Translation (Zh-En).** Following [30], we train our model on 1.25M sentence pairs with 27.9M Chinese

Table 10
Experimental Results of meSimp. It can be drawn from the results that meSimp could reduce the model to a smaller size, often around 10%, while maintaining the performance if not improving.

| Parsing | Iter | Size | Dev (%) | Test (%) |
|---|---|---|---|---|
| MLP (h=500) | 10 | 500 | 90.48 | 89.80 |
| meSimp (k=20, p=0.08) | 10 | **51 (10.2%)** | 90.23 | 90.11 (+0.31) |
| **POS** | Iter | Size | Dev (%) | Test (%) |
| LSTM (h=500) | 3 | 500 | 97.20 | 97.22 |
| meSimp (k=20, p=0.08) | 3 | **60 (12.0%)** | 97.19 | 97.25 (+0.03) |
| **MNIST** | Iter | Size | Dev (%) | Test (%) |
| MLP (h=500) | 13 | 500 | 98.72 | 98.20 |
| meSimp (k=160, p=0.10) | 14 | **154 (30.8%)** | 98.46 | 98.31 (+0.11) |

Table 11
meSimp: Adaptive Hidden Sizes. As we can see, meSimp can adaptively simplify the hidden layers to different and appropriate sizes.

| Parsing | Average | Hidden | |
|---|---|---|---|
| MLP | 500 | 500 | |
| meSimp(k=20, p=0.08) | 51 | 51 | |
| **POS** | Average | Forward | Backward |
| hline LSTM | 500 | 500 | 500 |
| meSimp (k=20, p=0.08) | 60 | 57 | 63 |
| **MNIST** | Average | First | Second |
| MLP | 500 | 500 | 500 |
| meSimp (k=160, p=0.10) | 154 | 149 | 159 |

Table 12
meSimp: Results Based on the Same $k$ and $h$ for Parsing. As we can see, the simplified model performs substantially better than the model of the same size trained from scratch. It shows that meSimp can efficiently select the best and the most robust features from big models.

| Parsing | Dev (%) | Test (%) |
|---|---|---|
| MLP (h=51) | 89.92 | 89.64 |
| meSimp (h=51) | **90.23** | **90.11** |
| Improvement | +0.31 | +0.47 |

words and 34.5M English words provided by LDC, and use NIST 2002 as the development set and NIST 2003-2006 as the test sets. The test score is reported as the average of the scores on individual test sets.

**English-Vietnamese Translation (En-Vi).** The training data is from the translated TED talks, containing 133K training sentence pairs provided by the IWSLT 2015 Evaluation Campaign [31]. The development set is TED tst2012 and the test set is TED tst2013.

The baseline model is based on LSTMs. Both the encoder and the decoder have three layers, and the encoder is bidirectional. The mini-batch size is 64. We use recurrent dropout [32], Luong-style attention [29], and beam search [33]. meProp is applied to each hidden layer. The evaluation metric is BLEU [34]. The results are summarized in Table 9. Machine translation is considered as an AI-complete task, which requires large-scale data and complex models to achieve competitive results. Nonetheless, meProp can still achieve good performance with only back propagating through 6.25% of the parameters in hidden layers, showing the potential and the generality of the approach.

## 3.4 Experimental Results of meSimp

Table 10 shows the model simplification results based on different models. In the table, *meSimp* means applying model simplification on top of meProp. $h = 500$ means that the dimension of the model's hidden layers is 500, $k = 20$ means that in back propagation we propagate top-20 elements, and $p = 0.08$ means that the neuron that is updated less than 8% times during a stage is dropped. *Size* is the average size of all the hidden layers.

As we can see, our method is capable of reducing the models to a relatively small size, while maintaining the performance if not improving. The hidden layers of the models are reduced by around 10x, 8x, and 3x for Parsing, POS, and MNIST respectively. The reason could be that the minimal effort update captures important and robust features, such that the simplified model is sufficient to represent the data, while without minimal effort update, the model of a similar size treats each feature equally at start, limiting its ability to learn from the data.

The results show that the simplifying method is effective in reducing the model size, thus bringing a substantial reduction of the computational cost of decoding in real-world task. More importantly, the accuracy of the original model is kept, or even more often improved. This means model simplifying could make it more probable to deploy a deep learning system to a computation constrained environment.

### 3.4.1 Analysis of meSimp

**Adaptive Hidden Size.** One of the advantages of meSimp is that it can automatically determine the hidden size of the model based on the data. The same effect is not achievable for most of the existing work ([22], [35], [36]), where the size needs be set in advance. As shown in Table 11, different hidden layers get different sizes with the same prune rate $p$. At the beginning, we conduct the experiments on a neural network with a single hidden layer, that is, Parsing, and we get promising result, as the model size is reduced to 10.2% of its original size. The result of Paring makes us wondering whether meSimp could also simplify deeper networks, so we continue to run experiments on different models. In the experiments of POS, there is a forward LSTM and a backward LSTM, which means it is often very deep in the temporal axis. As we expect, the forward and backward LSTMs indeed gets different dimensions, that is, 63 and 57 respectively. We further conduct experiments on an MLP with 2 hidden layers, and the result shows that the first hidden layer and the second hidden layer are again of different sizes, which confirms that meSimp could adaptively adjust the hidden layer size in a multi-layer setting.

**Comparing with Sames-Sized Small Models.** Another advantage of meSimp is that the simplified model could outperform a same-sized model trained from scratch. On the contrary, if the simplified model performs worse, it means the original model is much too big and the simplifying method fails to remove redundant neurons. It makes the whole simplification unnecessary, since simply training a small model is good enough. Fortunately, that is not the case, as shown in Table 12. We train baseline models of sizes the same as the sizes of simplified models, and our simplified models perform better than the models trained of similar sizes, especially on the Parsing task. The baseline is trained five times with different random seeds, and we choose the best to report. Despite that, our simplified model still performs much better. An intuitive explanation is that meSimp selects the most robust features from a large set of candidate features, while the smaller model trained from scratch lacks the diversity in its feature set.

Table 13
Experimental Results of meAct. It confirms our intuition that for each example only a small number of features are sufficient to make correct predictions, and only training the decisive features w.r.t. each example can be beneficial for the learning of the model.

| MNIST | Iter | Average Size | Test (%) |
|---|---|---|---|
| MLP (h=500) | 18 | 500 | 98.18 |
| meAct (p=0.004, e=10) | 20 | 99 | 98.42 (+0.24) |
| meAct (p=0.004, e=15) | 18 | 99 | 98.32 (+0.14) |

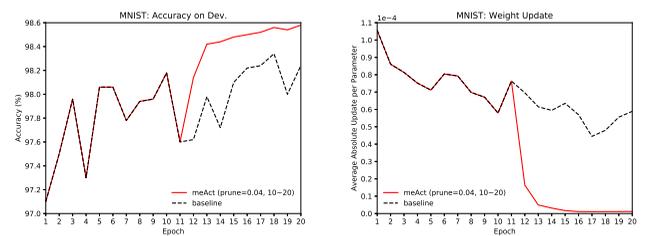

Figure 4. Change of accuracy (left), and average update of parameters (right). To isolate the impact of meAct, we fix the random seed. It will cause the identical learning process at different runs such that the lines coincide with each other during Epoch 1-10. As we can see, after meAct is applied, the accuracy rises, which indicates training focused on the most relevant neurons could reduce overfitting, and the update drops, which suggests in the later of normal training, most of the update is caused by fitting of the noises, making the already trained neurons constantly changing.

### 3.5 Why the Minimal Effort Technique Works?

From the back propagation simplification and model simplification results, we could see that approaches based on active paths, which are measured by the back propagation, are effective in reducing overfitting. One of our hypothesis is that for a neural network, for each example, only a small part of the neurons is needed to produce the correct results, and gradients are good identifiers to detect the decisive neurons. Too many neurons are harmful to the model, because the extra neurons may be trained to fit the noise in the training examples.

**meAct.** To examine the hypothesis, we design an new algorithm, which we call *meAct* (**m**inimal **e**ffort **act**ivation), which only activates the active path w.r.t. each example in the forward propagation. We first train the model normally to make it adapt to the data and obtain the accumulated absolute gradients for each example. Then, we use the gradients to choose the most active neurons for each example. Finally, we only train only train the most active neurons, so that the model make minimal effort in activation or forward propagation. The procedure is described in Algorithm 3.

---

**Algorithm 3** Minimal Effort Activation

**Input:** Pretrain epoch $e$, and activation threshold $p$
1: Train the model normally for $e$ epochs
2: **for all** neuron $i$ and example $j$ **do**
3:      Accumulate the absolute gradient $g_{ij}$
4:      Calculate the activation value $a_{ij}$
5: **end for**
6: **for all** example $j$ **do**    ▷ Select active neurons w.r.t. each example
7:      **if** $g_{ij} < p \times \sum_i g_{ij}$ **then**
8:          Mask neuron $i$ for example $j$ with $a_{ij}$
9:      **end if**
10: **end for**
11: Train the model with the non-masked neurons for each example

---

As the sparse activation is done in forward propagation, the back propagation is sparsified as well, because obviously, the deactivated neurons contribute none to the results, meaning their gradients are zeros, which requires no computation. Note that the method does not reduce the size of the model, and each example has its own active paths. During test, the forward propagation is done normally, as we wouldn't know the active paths of these unseen examples.

The results are summarized in Table 13. As we can see, for each example, fewer than 100 neurons are adequate to make good predictions and only train these neurons for that example can result in better results. To see how the accuracy improvement is acquired, we further investigate the change of the parameters during training. We calculate the average absolute parameter change at each update, and find that after only training the related neurons of each example, the change drops acutely.[2] We use the absolute of the update of a parameter, because we would like to see how much the parameters have been modified during the training process, not just the change between the start and the end.

Figure 4 illustrates the phenomenon. During minimal effort activation, the accuracy rises from the baseline, which shows that the accuracy could benefit from the training focused more on the decisive neurons. Besides, we can see that the update dropped sharply when meAct is applied, meaning the accuracy improvement is achieved by very little change of the parameters, while the update of normal training is still high, more than 5x of the update of meAct, suggesting that many update is redundant and unnecessary, which could be the result of the model trying to adapt to the noise in the data. As there should be no regular pattern in the noise, it requires more subtle update of all the parameters to fit the noise, which is much harder and often affects the training of the essential features, thus leading to a lower accuracy than our method which tries to focus only on the essential features for each example.

The results of this exploration attests to our initial hypothesis that for an example, only a few neurons are required, and minimal effort technique provides a simple yet effective way to train and extract the helpful neurons.

---

2. We use the accumulated absolute parameter change rather than the accumulated absolute gradient, because the Adam optimizer adaptively alters the scale of the gradient at each pass, making the parameter change quite different from the gradient.

## 4 RELATED WORK

### 4.1 Training Simplification

[37] proposed a direct adaptive method for fast learning, which performs a local adaptation of the weight update according to the behavior of the error function. [38] also proposed an adaptive acceleration strategy for back propagation. Dropout [21] was proposed to improve training speed and reduce the risk of overfitting. *Sparse coding* is a class of unsupervised methods for learning sets of overcomplete bases to represent data efficiently [39]. [40] proposed a sparse auto-encoder model for learning sparse overcomplete features. Our proposed method is quite different compared with those prior studies, in that our strategy is based on gradient information with a simple but effective strategy.

The sampled-output-loss methods [41] are limited to the softmax layer (output layer) and are only based on random sampling, while our method does not have those limitations. The sparsely-gated mixture-of-experts [42] only sparsifies the mixture-of-experts gated layer and it is limited to the specific setting of mixture-of-experts, while our method does not have those limitations.

There are also prior studies focusing on reducing the communication cost in *distributed systems* ([24], [43], [44], [45], [46]), by sparsifying the already computed gradients. For each instance, the gradient is calculated normally. Those settings are also different from ours.

### 4.2 Model Simplification

For model simplification, most of the existing methods prune the neural network based on the value of the parameters for smaller footprint and faster inference, starting from [47], [48] to more recent [35], [49], [50]. However, what they actually prune is the connection between the neurons, resulting in sparse matrix parameters requiring efficient sparse matrix operations. Different from them, our method prunes the neurons based on how many times a neuron is updated by the proposed sparsified back propagation, thus eliminating entire rows or columns in parameter matrices. The resulting model only needs the normal matrix operations, which are far more optimized than the sparse ones. [22] proposed to distill an expressive but cumbersome model into a smaller model by mimicking the target of the cumbersome model, while adjusting the temperature. They claim that their method can transfer the knowledge learned by the cumbersome model to the simpler one. And their method doesn't presume a specific model. However, the final model size should be preconfigured, while our method could adaptively learn the size of the model. [36] proposed a dense-sparse-dense model to first eliminate units with small absolute values, then reactivate the units, and re-train them, so that the model could achieve better performance. Their model does not really reduce the size of the model, as their purpose is to improve the performance.

Note that our work could adaptively choose the size of a layer in deep neural networks with a single simplifying configuration, as shown in Table 11. But most of the previous work, either the related hyper parameter controlling the final model size needs to be set for each layer ([35], [36]), or the model size needs to be set directly ([22]). Those methods will lead to trivial hyper-parameter tuning if different hidden layer sizes are pursued, as different layers, representing different levels of abstraction naturally should be of different sizes, which we do not know in advance. Our work eliminates the needs for that, as with the same hyper parameter, meSimp will automatically determine the size needed to represent the useful information for each hidden layer, thus leading to varied dimensions of the hidden layers.

### 4.3 Related Systems on the Tasks

The POS tagging task is a well-known benchmark task in natural language processing, with the accuracy reporting from 97.2% to 97.4% ([15], [51], [52], [53], [54], [55], [56], [57], [58], [59]). Our method achieves 97.31% (see Table 4).

For the transition-based dependency parsing task, existing approaches typically achieve the UAS score from 91.4 to 91.5 ([60], [61], [62], [63], [64], [65], [66], [67], [68], [69]). As one of the most popular transition-based parsers, MaltParser [64] has 91.5 UAS. [12] achieves 92.0 UAS using neural networks. Our method achieves 91.99 UAS (see Table 4).

For MNIST, the MLP based approaches can achieve 98–99% accuracy, often around 98.3% ([10], [26], [70]). Our method achieves 98.37% (see Table 5). With the help from convolutional layers and other techniques, the accuracy can be improved to over 99% ([71], [72]). Our method achieves 99.39% (see Table 8).

For machine translation tasks, using the same data splits, neural machine translation systems usually achieve BLEU scores from 35 to 39 on the Zh-En task ([73], [74], [75], [76], [77]), mostly around 38, and from 26 to 29 on the En-Vi task ([78], [79]). Ours reach 36.42 and 29.54, respectively (see Table 9). If the length normalization trick is used, our Zh-En model can be improved to 37.91. The En-Vi model reaches the highest BLEU score to our knowledge.

## 5 CONCLUSIONS

We propose a *minimal effort* back propagation technique to simplify the training (*meProp*), and to simplify the resulting model (*meSimp*).

The minimal effort technique adopts the top-$k$ selection based back propagation to determine the most relevant features, which leads to very sparsified gradients to compute for the given training sample. Experiments show that meProp can reduce the computational cost of back propagation by one to two orders of magnitude via updating only fewer than 5% parameters, and yet improve the model accuracy in most cases. We extend the techniques to convoluational neural networks and sequence-to-sequence models.

We further propose to remove the seldom updated parameters to simplify the resulting model for the purpose of reducing the computational cost of the decoding. Experiments reveal that the model size could be reduced to around one ninth of the original models, leading to around 9x computational cost reduction in decoding for two natural language processing tasks with improved accuracy. More importantly, meSimp could automatically decide the appropriate sizes for different hidden layers, alleviating the need for hyper-parameter tuning.

## ACKNOWLEDGMENTS

This work was supported in part by National Natural Science Foundation of China (No. 61673028), and an Okawa Research Grant (2016). This work is a substantial extension of the work presented at ICML 2017 [1].

**Xu Sun** is Associate Professor in Department of Computer Science, Peking University, since 2012. He got Ph.D. from The University of Tokyo (2010), M.S. from Peking University (2007), and B.E. from Huazhong Univ. of Sci. & Tech. (2004). From 2010 to 2012, he worked at The University of Tokyo, Cornell University, and The Hong Kong Polytechnic University as Research Fellow/Associate. His research focuses on natural language processing and machine learning, especially on structured natural language processing and structured learning. He has been Area Chair/Co-Chair of EMNLP, IJCNLP; Program committee member of ACL, IJCAI, AAAI, COLING, EMNLP, NAACL, PAKDD, ACML; Journal reviewer of IEEE TPAMI, Comput. Linguist., and so on.

**Xuancheng Ren** is a PhD candidate, supervised by Prof. Xu Sun, at School of Electronics Engineering and Computer Science, Peking University. He is currently a member of the MOE Key Laboratory of Computational Linguistics at Peking University. He received the degree of Bachelor of Science in Computer Science from Peking University in 2017. His work focuses on machine learning for natural language processing, especially on deep learning methods.

**Shuming Ma** received the Bachelor of Engineering degree from Peking University, China in 2016. He is now a graduate student in School of Electronics Engineering and Computer Science, Peking University. His current work concerns machine learning and natural language processing, particularly machine translation and text summarization. His papers are published on the top international conferences, including ACL, IJCAI, and ICML.

**Bingzhen Wei** is a master student at MOE Key Laboratory of Computational Linguistics, School of Electronics Engineering and Computer Science, Peking University. His main research interests focus on deep learning for natural language processing and understanding, particularly in machine translation.

**Wei Li** is currently a PhD candidate in the MOE Key Laboratory of Computational Linguistics, Peking University. From 2011 to 2015, he studied at the department of Computer Science, Peking University. His research areas include natural language processing, deep learning and the related. He has published several papers at conferences like COLING, NLPCC and CCL.

**Jingjing Xu** is a PhD candidate, supervised by Prof. Xu Sun. Her recent research interests include deep learning applied to natural language processing.

**Houfeng Wang** is a professor in the school of Electronic Engineering and Computer Science, Peking University (PKU). Now, he is the director of Institute of Computational Linguistics of PKU. His research interests are Natural Language Processing and Machine Learning.

**Yi Zhang** is a PhD candidate of Peking University, supervised by Prof. Xu Sun. Her research focuses on deep learning in natural language processing.